\documentclass[Afour,sagev,times]{sagej}

\usepackage{moreverb,url}
\usepackage[colorlinks,bookmarksopen,bookmarksnumbered,citecolor=red,urlcolor=red]{hyperref}

\usepackage[T1]{fontenc}
\usepackage{lmodern}
\usepackage{amsmath}
\usepackage{amssymb}
\usepackage{wrapfig}
\usepackage{graphicx}
\usepackage{algorithm2e}
\usepackage{eqparbox}
\usepackage{mdwmath}
\usepackage{mdwtab}
\usepackage{array}
\usepackage{authblk}

\usepackage{hyperref}
\usepackage{units}
\usepackage{multirow}

\usepackage{subfigure}

\usepackage{setspace}

\graphicspath{
	{figs/}
	{evaluation/}
	{anim/}
}

\definecolor{MyBrickRed}{cmyk}{0,0.89,0.94,0.28}

\usepackage[colorinlistoftodos, textwidth=3.7cm]{todonotes}
\renewcommand{\vec}[1]{\mathbf{#1}}
\newcommand\norm[1]{\left\lVert#1\right\rVert}

\newcommand\BibTeX{{\rmfamily B\kern-.05em \textsc{i\kern-.025em b}\kern-.08em
T\kern-.1667em\lower.7ex\hbox{E}\kern-.125emX}}

\def\volumeyear{2021}

\begin{document}

\runninghead{
}

\title{LVD-NMPC: A Learning-based Vision Dynamics Approach to Nonlinear Model Predictive Control for Autonomous Vehicles}

\author{
    Sorin~Grigorescu, Cosmin~Ginerica, Mihai~Zaha, Gigel~Macesanu, Bogdan~Trasnea
}

\affiliation{
    The authors are with RovisLab: Robotics, Vision and Control Laboratory (\url{www.rovislab.com}), Transilvania University of Brasov, Romania and Elektrobit Automotive (\url{www.elektrobit.com}).
}

\corrauth{
    Sorin Grigorescu, 
    Institute of Automation
    Transilvania University of Brasov
    Mihai Viteazu 5, Corp V, et. 3
    500174 Brasov, Romania
}

\email{
    s.grigorescu@unitbv.ro
}

\begin{abstract}
In this paper, we introduce a learning-based vision dynamics approach to nonlinear model predictive control for autonomous vehicles, coined LVD-NMPC. LVD-NMPC uses an a-priori process model and a learned vision dynamics model used to calculate the dynamics of the driving scene, the controlled system's desired state trajectory and the weighting gains of the quadratic cost function optimized by a constrained predictive controller. The vision system is defined as a deep neural network designed to estimate the dynamics of the images scene. The input is based on historic sequences of sensory observations and vehicle states, integrated by an Augmented Memory component. Deep Q-Learning is used to train the deep network, which once trained can be used to also calculate the desired trajectory of the vehicle. We evaluate LVD-NMPC against a baseline Dynamic Window Approach (DWA) path planning executed using standard NMPC, as well as against the PilotNet neural network. Performance is measured in our simulation environment GridSim, on a real-world 1:8 scaled model car, as well as on a real size autonomous test vehicle and the nuScenes computer vision dataset.
\end{abstract}


\maketitle

\section{Introduction}
\label{sec:introduction}

Research in the area of autonomous driving has been boosted in the last decade both by academia and industry. Autonomous vehicles are intelligent agents equipped with driving functions designed to understand their surroundings and derive control actions. As shown in the deep learning for autonomous driving survey of Grigorescu et al.~\cite{Grigorescu_JFR_2019}, the driving functions are traditionally implemented as perception-planning-action pipelines. 
Recently, approaches based on End2End learning from Bojarski et al.~\cite{NVIDIA_end2end_Learning} and Pan et al.~\cite{Pan_IJRR_19}, or the Deep Reinforcement Learning (DRL) shown by Kendall et al.~\cite{Kendall2018} have also been proposed, although mostly as research prototypes.

In a modular perception-planning-action system, visual perception is most of the times decoupled from low-level control. A tighter coupling of perception and control was researched in the field of robotic manipulation with the concept of \textit{visual servoing}, as in the case of the manipulation fault detector og Gu et al.~\cite{9219236}. However, this is not the case in autonomous vehicles, were intrinsic dependencies between the different modules of the driving functions are not taken into account.

\begin{figure}
	\centering
	\begin{center}
		\includegraphics[scale=0.98]{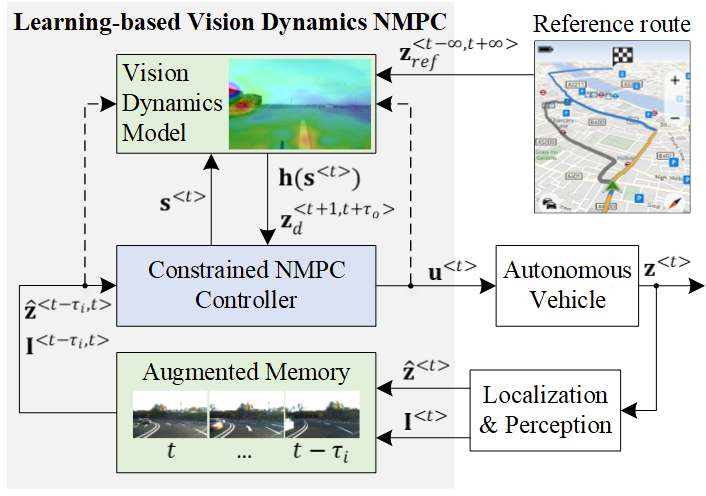}
		\caption{\textbf{LVD-NMPC: Learning-based Vision Dynamics Nonlinear Model Predictive Control for autonomous vehicles}. The dotted lines illustrate the data flow used during training.}
        \label{fig:block_diagram}
	\end{center}
\end{figure}

This work is a contribution in the area of vision dynamics and control, where the proposed \textit{Learning-based Vision Dynamics Nonlinear Model Predictive Control} (LVD-NMPC) algorithm is used for controlling autonomous vehicles. An introduction of the vision dynamics concept in learning control can be found in the work of Grigorescu~\cite{GRIGORESCU2021243}. The block diagram of LVD-NMPC is shown in Fig.~\ref{fig:block_diagram}, where the main components are a vision dynamics model defined as a Deep Neural Network (DNN) and a constrained non-linear model predictive controller (NMPC) which receives input from the vision model. The model, trained using the Q-learning algorithm, calculates desired state trajectories and the tuning gains of the NMPC.

Synergies between data driven and classical control methods have been considered for imitation learning, where steering and acceleration control signals have been calculated in an End2End manner, as by Pan et al.~\cite{Pan_IJRR_19}. Their approach is designed for driving environments with predefined boundaries, without any obstacles present on the driving track.

As shown by Grigorescu et al.~\cite{Grigorescu_JFR_2019}, traditional decoupled visual perception systems use visual localization to estimate the pose of the ego-vehicle relative to a reference trajectory, together with obstacle detection. The information is further used by a path and behavioral planner to determine a safe driving trajectory, which is executed by the motion controller. In our work, we improve the traditional visual approach by replacing the classical perception-planning pipeline with a learned vision dynamics model. The model is used to calculate a safe driving	 trajectory, as well as to estimate the optimal tuning gains of the NMPC's quadratic cost function. Our formulation exploits the advantages of model-based control with the prediction capabilities of deep learning methods, enabling us to encapsulate the vision dynamics within the layers of a DNN. 
The key contributions of the paper are:

\begin{enumerate}
    \item the autonomous vehicle LVD-NMPC controller, based on an a-priori process model and a Vision Dynamics model;
    \item a deep neural network architecture acting as a nonlinear vision dynamics approximator used to estimate optimal desired state trajectories and the NMPC's tuning gains;
    \item a method for training the LVD-NMPC controller, based on imitation learning and the Q-Learning training approach.
\end{enumerate}


The rest of the paper is organized as follows. The related work is covered in the next section. The methodology of LVD-NMPC is given in \textit{Vision Dynamics Model Learning and
Control System}, followed by the experimental results. Finally, the paper is summarized in the conclusions section.

\section{Related Work}
\label{sec:related_work}

\begin{figure*}
	\centering
	\begin{center}
		\includegraphics[scale=1.07]{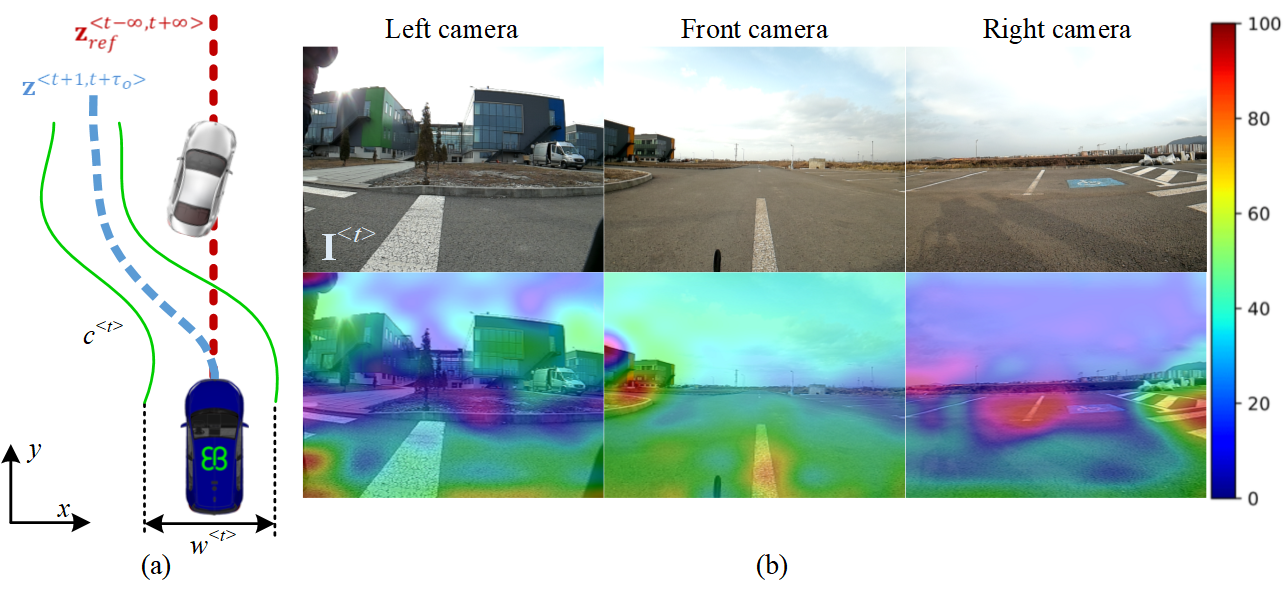}
		\caption{\textbf{Autonomous driving problem definition.} Given the vehicle's state, the global route (red line)and a set of sensory measurements, the goal is to calculate a safe path for tracking (blue line) over the control horizon $[t+1, t+\tau_o]$. The vision dynamics model estimates the curvature $c^{<t>}$ and width $w^{<t>}$ of the road. (a) Illustration of obstacle avoidance. (b) Observation $\vec{I}^{<t>}$ (top images) and activation maps (bottom images) within the vision dynamics model. The chromatic scale shows the contribution of the neurons in the activation maps at pixel level, where red corresponds to a high contribution and blue to almost no contribution (best viewed in color).}
        \label{fig:problem_description}
	\end{center}
\end{figure*}

Recent years have witnessed a growing trend in applying deep learning techniques to autonomous driving, especially in the areas of \textit{End2End} learning, as in the methods proposed by Pan et al.~\cite{Pan_IJRR_19}, Fan et al.~\cite{SNE_RoadSeg} and Bojarski et al.~\cite{NVIDIA_end2end_Learning}, as well as in \textit{Deep Reinforcement Learning} (DRL). Relevant algorithms for self-driving based on DLR can be found in the works of Kiran et al.~\cite{9351818}, Kendal et al.~\cite{Kendall2018} and Wulfmeier et al.~\cite{WulfmeierWP16}. Flavors of machine learning techniques have also been encountered in more traditional control approaches, such as \textit{Nonlinear Model Predictive Control} (NMPC), such as the uncertainty aware NMPC of Lucia and Karg~\cite{LUCIA2018511} and the \textit{Learning Controllers} of Ostafew et al.~\cite{ostafew-ijrr16} and McKinnon and Schoellig~\cite{mckinnon-ral19}.

\textit{End2end learning}, as described by Amini et al.~\cite{8957584}, directly maps raw input data to control signals. The approach in LVD-NMPC is similar to the one considered by Pan et al.~\cite{Pan_IJRR_19}. Compared with our method, their DNN policy is trained for agile driving on a predefined obstacle-free track. This approach limits the applicability of their system to autonomous driving, since a self-driving car has to navigate roads with dynamics obstacles and undefined lane boundaries. An end2end neural motion planner has been proposed by Zeng et al.~\cite{Zeng_2019_CVPR}, while Fan et al.~\cite{SNE_RoadSeg} designed an End2End learning system to predict the drivable surface. The work considers no obstacle detection and avoidance, improving solely the perception system, without tackling the intrinsic dependencies between perception and low-level vehicle control.

\textit{Deep Reinforcement Learning} (DRL) is a type of machine learning algorithm where agents are taught actions by interacting with their environment. An extensive review of DRL for autonomous driving has been published by Kiran et al.~\cite{9351818}. The main challenge with DRL for real-world physical systems is that the agent, in our case a self-driving car, learns by exploring its environment. A solution here is provided by \textit{Inverse Reinforcement Learning} (IRL), which is an imitation learning method for solving Markov Decision Processes (MDPs). Wulfmeier et al.~\cite{WulfmeierWP16} have extended the Maximum Entropy IRL with a convolutional DNN for learning a navigation cost map in urban environments. However, such methods usually do not take into consideration the vehicle's state and the feedback loop required for low-level control.

\textit{Nonlinear Model Predictive Control} (NMPC), as presented by Garcia et al.~\cite{GARCIA_Automatica_1989}, is a control strategy which computes control actions by solving an optimization problem tailored around a nonlinear dynamic system model. In the last decades, it has been successfully applied to autonomous driving applications, both in research as well as in the automotive industry. NMPC controllers have been proposed by Nascimento et al.~\cite{8713577}and~\cite{Nascimento_2018} for trajectory tracking in nonholonomic mobile robots. In order to deal with uncertainties, learning based approaches to model predictive control have been used by Lucia and Karg~\cite{LUCIA2018511}, as well as by Gango et. all~\cite{GANGO2019152} for approximating an explicit NMPC system.

\textit{Learning Controllers}. Traditional feedback controllers, such as NMPC, make use of an \textit{a-priori} model composed of fixed parameters. Unlike controllers with fixed parameters, learning controllers make use of training information to learn their models over time. In previous works, learning controllers have been introduced based on simple function approximators, such as the Gaussian Process (GP) modeling in the work of Ostafew et al.~\cite{ostafew-ijrr16} or bayesian regression algorithm of Mckinnon and Schoellig~\cite{mckinnon-ral19}.

In the light of the current approaches and their limitations, the LVD-NMPC method is proposed for encapsulating the driving scene's dynamics within a vision dynamics model, which adapts a constrained NMPC for executing the desired vehicle state trajectory.

\section{Vision Dynamics Model Learning and Control System}
\label{sec:method}

\subsection{Problem Definition}

Fig.~\ref{fig:problem_description} shows a simple illustration of the autonomous driving problem. Given past vehicle states $\vec{z}^{<t-\tau_i, t>}$, a sequence of observations $\vec{I}^{<t-\tau_i, t>}$ and a global reference route for tracking $\vec{z}^{<t-\infty, t+\infty>}_{ref}$, the task is to calculate the control signals $\vec{u}^{<t+1>}_{opt}$ for time $t+1$, such that the self-driving car tracks a safe trajectory $\vec{z}^{<t+1, t+\tau_o>}$.
Considering a discrete sampling time $t$, $\tau_i$ and $\tau_o$ are past and future temporal horizons, respectively.

The reference trajectory $\vec{z}_{ref}$ represents the global route which should be followed by the vehicle, from its start position to destination. It can be given as a set of waypoints (e.g. GPS coordinates). Since $\vec{z}_{ref}$ is a global trajectory, it is considered to vary in the $[t-\infty, t+\infty]$ interval. $\vec{z}_{ref}$ is different from the desired trajectory $\vec{z}^{<t+1, t+\tau_o>}_d$, the later one  being calculated for the interval $[t+1, t+\tau_o]$. $\vec{z}_d$ is executed by the NMPC controller and must take into account the drivable area and the obstacles present in the scene.

The vehicle is modeled based on the kinematic bicycle model of a robot described by Paden et al.~\cite{PadenCYYF16}, with position state $\vec{z}^{<t>} = (x^{<t>}, y^{<t>}, \rho^{<t>})$ and no-slip assumptions. $x$, $y$ and $\rho$ represent the position and heading of the vehicle in the 2D driving plane, respectively. We apply the longitudinal velocity and the steering angle as control actions: $\vec{u}^{<t>} = (v^{<t>}_{cmd}, \omega^{<t>}_{cmd})$, where $\omega^{<t>}_{cmd}$ is the angle of the current velocity of the center of mass with respect to the longitudinal axis of the car, as proposed by Borrelli et al.~\cite{Borrelli_Kinematic_Dynamic_Models}.

The driving scene is modeled as the vision dynamic state $(c^{<t>}, w^{<t>})$, where $c$ and $w$ are the curvature and traversable width of the road. A high $w$ corresponds to relaxed constraints when estimating the desired trajectory, while a low $w$ represents a path that has to be followed precisely in order to satisfy safety constraints. We consider a normalized value for the traversable width $w \in [0,1]$, where $1$ is the maximum road width and $0$ signals a non-traversable area, in which case the vehicle has to stop.

When acquiring training samples, the following quantities are stored as sequence data: the historic position states $\vec{z}^{<t-\tau_i, t>}$, the sensory information acquired from a front facing camera $\vec{I}^{<t-\tau_i, t>}$, the global reference trajectory $\vec{z}^{<t-\tau_i, t+\tau_o>}_{ref}$ and the control actions $\vec{u}^{<t-\tau_i, t>}$ recorded from a human driver. For practical reasons, the global reference trajectory is stored at sampling time $t$ over the finite horizon $[t - \tau_i, t + \tau_o]$. A single image corresponds to an observation instance $\vec{I}^{<t>}$, while a continuous sequence of images is denoted as $\vec{I}^{<t-\tau_i, t>}$.

\subsection{Control System Design}
\label{sec:rcnmpc}

The block diagram of LVD-NMPC is shown in Fig.~\ref{fig:block_diagram}. Consider the following nonlinear, state-space system:

\begin{equation}
	\vec{z}^{<t+1>} = \vec{f}_{true} (\vec{z}^{<t>}, \vec{u}^{<t>}),
\end{equation}

\noindent where $\vec{z}^{<t>} \sim \mathcal{N} (\vec{\bar{z}}^{<t>}, \sigma^2_f)$ is the observable state, $\vec{z}^{<t>} \in \mathbb{R}^n$, $\vec{\bar{z}}^{<t>}$ is the mean state and $\vec{u}^{<t>} \in \mathbb{R}^m$ is the control input at discrete time $t$. We assume that each state measurement is corrupted by zero-mean additive Gaussian noise with variance $\sigma^2_f$. $\vec{f}_{true}$ is approximated as a sum between an \textit{a-priori} model and an experience-based vision dynamics model:

\begin{equation}
	\vec{z}^{<t+1>} = \underset{\text{\textit{a-priori} model}}{\vec{f} (\vec{z}^{<t>}, \vec{u}^{<t>})} + \underset{\text{vision dynamics model}}{\vec{h} (\vec{s}^{<t-\tau_i, t>})},
	\label{eq:true_system_model}
\end{equation}

\noindent where $\vec{s}^{<t>} \in \mathbb{R}^p$ represent environmental dependencies.

This dependencies are composed of the system's and environment's states at time $t$, defined as:

\begin{equation}
    \vec{s}^{<t>} = (\vec{z}^{<t-\tau_i, t>}, \vec{I}^{<t-\tau_i, t>}),
\end{equation}

\noindent where $\vec{s}^{<t>}$ represents the set of historic dependencies integrated along time interval $[t-\tau_i, t]$ by the so-called \textit{Augmented Memory} component.

The models $\vec{f}(\cdot)$ and $\vec{h}(\cdot)$ are nonlinear process models. $\vec{f}(\cdot)$ is a known process model, representing our knowledge of $\vec{f}_{true}(\cdot)$, while $\vec{h}(\cdot)$ is the learned vision dynamics model. With the sampling time defined as $\delta t$, the nominal process model employed by LVD-NMPC is:

\begin{equation}
	\vec{f} (\vec{z}^{<t>}, \vec{u}^{<t>}) = \vec{z}^{<t>} + 
		\delta t
		\begin{bmatrix}
			\cos (\rho^{<t>} + \omega^{<t>}_{cmd}) \\
			\sin (\rho^{<t>} + \omega^{<t>}_{cmd}) \\
			\frac{1}{L} \sin (\omega^{<t>}_{cmd}))
		\end{bmatrix}
		v^{<t>}_{cmd}
	\label{eq:nominal_process_model}
\end{equation}

\noindent where $L$ is the length between the front and the rear wheel.

We distinguish between the given reference trajectory $\vec{z}^{<t-\infty, t+\infty>}_{ref}$, which, from a control perspective, is practically infinite, and a desired state trajectory $\vec{z}^{<t+1, t+\tau_o>}_d$, calculated over a finite prediction horizon $\tau_o$. The optimal future states are computed by a constrained NMPC controller, based on a desired state trajectory $\vec{z}^{<t+1, t+\tau_o>}_d$, estimated by the vision dynamics model.

The vision dynamics model learns to predict the driving scene's dynamics $(c^{<t>}, w^{<t>}) \sim \mathcal{N} (\bar{c}^{<t>}, \bar{w}^{<t>}, \sigma^2_h)$ from historic states integrated by the augmented memory component. The visual measurements are corrupted by zero-mean Gaussian noise with variance $\sigma^2_h$. When queried, it returns a prediction for $c^{<t>}$ and $w^{<t>}$, given a sequence of historic states. The $\vec{h}(\cdot)$ model in Eq.~\ref{eq:true_system_model} is calculated by multiplying $c$ and $w$ with three weighting constants. $c$ is used to correct the heading, while $w$ adapts the position of the vehicle in the 2D driving plane.

The scene's dynamics are used to calculate the desired trajectory $\vec{z}^{<t+1, t+\tau_o>}_d$, taking into account the global reference route $\vec{z}^{<t-\infty, t+\infty>}_{ref}$. $\vec{z}^{<t+1, t+\tau_o>}_d$ can be interpreted as the quantitative deviation of the vehicle from the reference route:

\begin{equation}
	\vec{z}^{<t+1, t+\tau_o>}_d = \vec{z}^{<t+1, t+\tau_o>}_{ref} + \vec{z}^{<t+1, t+\tau_o>}_c,
\end{equation}

\noindent where $\vec{z}^{<t+1, t+\tau_o>}_c$ is the vision dynamics estimation for the optimal trajectory of the vehicle's path, computed in $xy$-coordinates relative to the vehicle's pose. Once $\vec{h}(\cdot)$ has been queried, the coordinates of the predicted path are sampled over $[t+1, t+\tau_o]$ based on $(c^{<t>}, w^{<t>})$, as:

\begin{equation}
	y^{<t+i>} = w^{<t>} - \rho \cdot^{<t>} (x^{<t+i>} - l) + 0.5 \cdot c^{<t>} \cdot {x^{<t+i>}}^2,
	\label{eq:road_projection}
\end{equation}

\noindent where $(t+i) \in [t+1, t+\tau_o]$ and $l$ is a lookahead distance calculated for a sampling time of $3s$.

As detailed in the Learning a Vision Dynamics Model section, a DNN is utilized to encode $\vec{h}(\cdot)$ and on top of $\vec{h}(\cdot)$, we define a quadratic cost function to be optimized by the constrained NMPC over discrete time interval $[t+1, t+\tau_o]$:

\begin{equation}
	J (\vec{z}, \vec{u}) = (\vec{z}_d - \vec{z})^T \vec{Q} (\vec{z}_d - \vec{z}) + \vec{u}^T \vec{R} \vec{u},
	\label{eq:nmpc_cost_function}
\end{equation}

\noindent where $\vec{Q} \in \mathbb{R}^{\tau_o n \times \tau_o n}$ is a positive semi-definite scalar state weight matrix, $\vec{R} \in \mathbb{R}^{\tau_o M \times \tau_o M}$ is a positive definite scalar input weight matrix, $\vec{z}^{<t+1, t+\tau_o>}_d$ is a sequence of desired states estimated by the vision dynamics model:

\begin{equation}
    \vec{z}^{<t+1, t+\tau_o>}_d = [\vec{z}^{<t+1>}_d, ..., \vec{z}^{<t+\tau_o>}_d],
\end{equation}

\noindent where $\vec{z}^{<t+1, t+\tau_o>}$ is the sequence of predicted states:

\begin{equation}
    \vec{z}^{<t+1, t+\tau_o>} = [\vec{z}^{<t+1>}, ..., \vec{z}^{<t+\tau_o>}],
\end{equation}

\noindent and $\vec{u}^{<t, t+\tau_o -1>}$ is the control input sequence:

\begin{equation}
    \vec{u}^{<t, t+\tau_o -1>} = [\vec{u}^{<t>}, ..., \vec{u}^{<t+\tau_o -1>}].
\end{equation}

The curvature $c^{<t>}$ is calculated using a polynomial interpolation of the trajectory points in $\vec{z}^{<t+1, t+\tau_o>}_d$, while $w^{<t>}$ is proportional to the velocity of the agent (maximum velocity corresponds to $w^{<t>} = 1$, while no motion is represented as $w^{<t>} = 0$). Eq.~\ref{eq:road_projection} is used to convert between trajectories and $(c^{<t>}, w^{<t>})$. At training time, $\vec{z}_d$ is given as human driven trajectories.

The traversable width $w^{<t>} \in [0,1]$ is used to automatically determine the ratios between the state and input weight matrices in Eq.~\ref{eq:nmpc_cost_function}. The operation is performed by weighting the coefficients of the main diagonals in $R$ and $Q$ based on $w$, as:

\begin{equation}
    diag(Q) = w^{<t>},
\end{equation}

\begin{equation}
    diag(R) = 1 - w^{<t>}.
\end{equation}

In this way, more aggressive control actions can be chosen when the road has high traversability, indicated by a high value of $w$.


The constrained NMPC objective is to find a set of control actions that optimize the vehicle's motion over a given time horizon $\tau_o$, while satisfying a set of hard and soft constraints:

\begin{subequations}
	\begin{equation}
		(\vec{z}^{<t+1>}_{opt}, \vec{u}^{<t+1>}_{opt}) = \underset{\vec{z}, \vec{u}}{\arg\min} \text{ } J \text{ } (\vec{z}^{<t+1, t+\tau_o>}, \vec{u}^{<t+1, t+\tau_o>})
        \label{eq:nmpc_problem}
	\end{equation}
	\begin{flalign}
		& \text{such that } \vec{z}^{<0>} = \vec{z}^{<t>} &
	\end{flalign}
	\begin{flalign}
		&
		\begin{split}
			\vec{z}^{<t+i+1>} =	& \vec{f} (\vec{z}^{<t>}, \vec{u}^{<t>}) + \vec{h} (\vec{s}^{<t-\tau_i, t>}),
		\end{split}
		&&
	\end{flalign}
	\begin{flalign}
		&
		\begin{split}
			\vec{e}^{<t+i>}_{\min} \leq \vec{e}^{<t+i>} \leq \vec{e}^{<t+i>}_{\max}, 
		\end{split}
		&&
	\end{flalign}
	\begin{flalign}
		&
		\begin{split}
			\vec{u}^{<t+i>}_{\min} \leq \vec{u}^{<t+i>} \leq \vec{u}^{<t+i>}_{\max}, 
		\end{split}
		&&
	\end{flalign}
	\begin{flalign}
		&
		\begin{split}
			\vec{\dot{u}}^{<t+i>}_{\min} \leq \frac{\vec{\dot{u}}^{<t+i>} - \vec{\dot{u}}^{<t+i-1>}}{\Delta t} \leq \vec{\dot{u}}^{<t+i>}_{\max}, 
		\end{split}
		&&
	\end{flalign}
\end{subequations}

\noindent where $i = 0, 1, ..., \tau_o - 1$, $\vec{z}^{<0>}$ is the initial state and $\Delta t$ is the sampling time of the controller. $\vec{e}^{<t+i>} = \vec{z}^{t+i}_d - \vec{z}^{t+i}$ is the cross-track error, $\vec{e}^{<t+i>}_{min}$ and $\vec{e}^{<t+i>}_{max}$ are the lower and upper tracking bounds, respectively. Additionally, we consider $\vec{u}^{<t+i>}_{\min}$, $\vec{\dot{u}}^{<t+i>}_{\min}$ and $\vec{u}^{<t+i>}_{\max}$, $\vec{\dot{u}}^{<t+i>}_{\max}$ as lower and upper constraint bounds for the actuator and actuator rate of change, respectively. The DL-NMPC-RSD controller implements:

\begin{equation}
	\vec{u}^{<t>} = \vec{u}^{<t+1>}_{opt},
\end{equation}

\noindent at each iteration $t$.

We leverage on the quadratic cost function from Eq.~\ref{eq:nmpc_cost_function} and solve the nonlinear optimization problem described above using the Broyden-Fletcher-Goldfarb-Shanno algorithm of Fletcher~\cite{Flet87}. The optimization problem from Eq.~\ref{eq:nmpc_problem} has been solved in real-time using the C++ version of the open-source Automatic Control and Dynamic Optimization (ACADO) toolkit of Houska et al.\cite{Houska2011a, Houska2011b}.

\subsection{Learning a Vision Dynamics Model}
\label{sec:vision_dynamics_model}

\begin{figure*}
	\centering
	\begin{center}
		\includegraphics[scale=0.92]{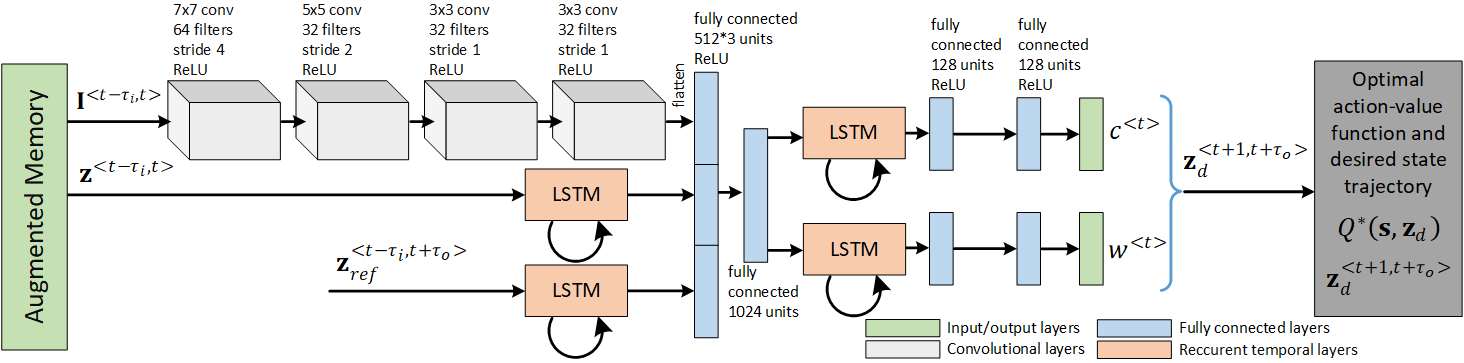}
		\caption{\textbf{Vision Dynamics model implemented as a deep neural network.} The training data consists of observation sequences, historic system states and reference state trajectories. A convolutional neural network firstly processes the observations data stream. Secondly, separate LSTM branches are responsible for calculating the road's curvature and width, which are then used to obtain the desired path.}
        \label{fig:neural_network_diagram}
	\end{center}
\end{figure*}

The role of the vision dynamics model is to estimate the scene's dynamics $(c^{<t>}, w^{<t>})$ using the temporal information stored in the Augmented Memory component and the global reference trajectory $\vec{z}^{<t-\infty, t+\infty>}_{ref}$. For practical reasons, we consider the reference trajectory to vary within the finite time interval $[t-\tau_i, t+\tau_o]$. The model is encoded by combining a Convolutional DNN with the robust temporal predictions of two Long Short-Term Memory (LSTM) networks, one for each element in $(c^{<t>}, w^{<t>})$.

Although the model could learn local state sequences directly, as in the previous NeuroTrajectory work of Grigorescu et al.~\cite{Grigorescu_RAL_2019}, we have chosen to learn the vision dynamics model $(c^{<t>}, w^{<t>})$, which can be used both for state prediction in the form of ego-vehicle poses, as well as for tuning the NMPC's quadratic cost function.

Given a sequence of temporal observations $\vec{I}^{<t-\tau_i, t>}: \mathbb{R}^w \times \tau_i \rightarrow \mathbb{R}^w \times \tau_o$, the system's state $\vec{z}^{<t>} \in \mathbb{R}^n$ in $\vec{I}^{<t>}$ and the reference set-points $\vec{z}^{<t+\tau_o>}_{ref} \in \mathbb{R}^n$ in observation space at time $t$, the task is to learn $(c^{<t>}, w^{<t>})$ in order to navigate from state $\vec{z}^{<t>}$ to destination state $\vec{z}^{<t+\tau_o>}$.

In reinforcement learning terminology, the autonomous driving problem can be described as a \textit{Partially Observable Markov Decision Process} (POMDP):

\begin{equation}
    M := (I, S, Z_d, \mathcal{T}, R, \gamma),
\end{equation}

\noindent where:

\begin{itemize}
    \item $I$ represents the sensory measurements;
    \item $S$ is a finite set of states;
    \item $Z_d$ is a set of trajectory sequences, used by the vehicle to navigate the driving environment measured via $\vec{I}^{<t-\tau_i, t>}$;
    \item $\mathcal{T}: S \times Z_d \times S \rightarrow [0, 1]$ is a stochastic transition function;
    \item $R: S \times A \times S \rightarrow \mathbb{R}$ is a scalar reward controlling the computation of $\vec{z}^{<t+1, t+\tau_o>}_d$;
    \item $\gamma$ is a discount factor regulating the importance of future versus immediate rewards.
\end{itemize}

The training objective is to find the desired trajectory that maximizes the associated cumulative future reward. We define the optimal action-value function $Q^*(\cdot, \cdot)$ for computing future discounted rewards for the starting $\vec{s}^{<t>}$ and control commands $\vec{u}^{<t+1, t+\tau_o>}$ for the state trajectory $\vec{z}^{<t+1, t+\tau_o>}_d$:

\begin{equation}
	Q^* (\vec{s}, \vec{z}_d) = \underset{\pi}{\max} \mathbb{E} \text{ } [R^{<\hat{t}>} | \vec{s}^{<\hat{t}>} = \vec{s}, \text{ } \vec{z}^{<\hat{t}+1, \hat{t}+\tau_o>}_d = \vec{z}_d, \text{ } \pi],
	\label{eq:optimal_action_value_function}
\end{equation}

\noindent where $\pi$ is an action, also known as policy, encapsulating a probability density function over a set of possible actions that can take place in a given state. For computing $Q^* (\vec{s}, \vec{z}_d)$, we propose the DNN from Fig.~\ref{fig:neural_network_diagram}, where sequences of consecutive images are processed by a set of convolutional layers, before being fed to two LSTM branches.

Our deep network is processing sequences of continuous temporal observations from the \textit{Augmented Memory} component. The Augmented memory acts as a buffer where the observations $\vec{I}^{<t-\tau_i, t>}$ and vehicle states $\vec{z}^{<t-\tau_i, t>}$ are synchronized and stored for the past temporal interval $[t-\tau_i, t]$, where $t$ is the current time and $\tau_i$ is the maximum amount of time for which we store observations and states.

The architecture of our DNN is mainly based on convolutional, recurrent and fully connected layers, illustrated in Fig.~\ref{fig:neural_network_diagram} in gray, orange and blue, respectively. The sequence of four convolutional layers are used for encoding the visual input into a latent one-dimensional intermediate representation that can be fed to the subsequent recurrent layers. In particular, the visual input is firstly passed through $64$ convolutional filters of size $7 \times 7$, followed by $32$ filters of size $5 \times 5$ and two blocks of $32$ filters, both having a $3 \times 3$ size. The recurrent layers have been implemented as Long-Short Term Memory (LSTM) networks, where the input is represented by the sequences stored in the Augmented Memory. The first two LSTMs are used to encode the sequence of vehicle states $\vec{z}^{<t, t-\tau_i>}$ and given reference trajectory $\vec{z}^{<t-\tau_i, t+\tau_o>}_{ref}$. The outputs are concatenated as a flatten extension of the one-dimensional latent representation. The complete latent space is composed of $1.536$ neurons, all activated using the Rectified Linear Unit (ReLU) activation function. For improving convergence during training, the intermediate representation reduced to $1.024$ neuron units using fully an additional connected layer. The network then branches into two heads responsible for estimating the current curvature $c^{<t>}$ and width $w^{<t>}$ of the road, respectively. Both branches use an LSTM and two sequential fully connected layers of $128$ neurons each, also activated using the ReLU function. Finally, the obtained curvature and road width are used to project the desired future trajectory of the vehicle $\vec{z}^{<t+1,t+\tau_o>}_d$, which in turn is used for calculating the optimal action-value function in Eq.~\ref{eq:optimal_action_value_function}.

One of the biggest challenges of using data-driven techniques for control is the so-called "DaGGer effect" described by Pan et al.~\cite{Pan_IJRR_19}, which is a significant drop in performance when the training and testing trajectories are significantly different. One measure to cope with this phenomenon is to ensure that sufficient data is provided at training time, thus increasing the generalization capabilities of the neural network. The "DaGGer effect" is also the main reason why a deep Q-learning approach, which uses the reward function to explore different trajectories, is preferred over standard supervised imitation learning. In the next section, it is shown that a high generalization can be achieved, demonstrating that LVD-NMPC can safely navigate the driving environment, even if the encountered obstacles were not given at training time. 


\section{Experiments}
\label{sec:experiments}

The performance of LVD-NMPC was benchmarked against a baseline non-learning approach, coined DWA-NMPC, as well as against PilotNet of Bojarski et al.~\cite{NVIDIA_end2end_Learning}. DWA-NMPC uses the Dynamic Window Approach (DWA) proposed by Fox et al.~\cite{Fox_Dynamic_Window_Approach_1997, dwa} for path planning and a constraint NMPC for motion control, relying for perception on the YoloV3 algorithm of Redmon and Farhadi~\cite{redmon2018yolov3}.

\begin{figure*}
	\centering
	\begin{center}
		\includegraphics[scale=1.3]{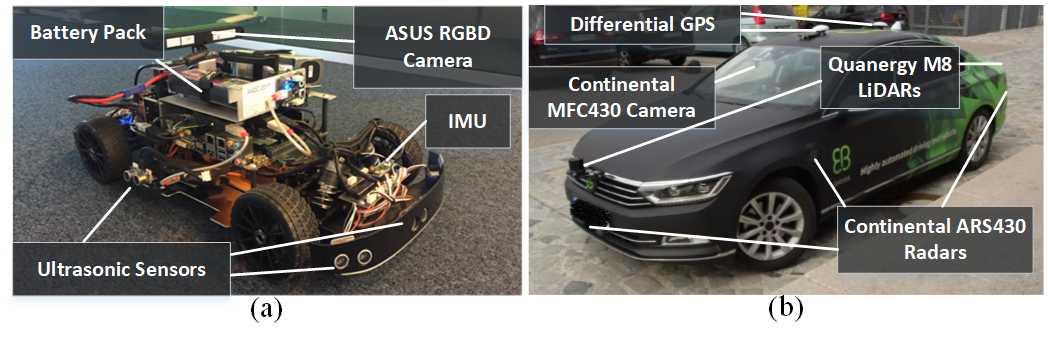}
		\caption{\textbf{Test vehicles used for data acquisition and testing.} (a) Audi 1:8 scaled model car. (b) Real-sized VW Passat autonomous test vehicle.}
        \label{fig:eb_car}
	\end{center}
\end{figure*}

LVD-NMPC has been tested on three different environments: \textit{I}) in the GridSim simulator, \textit{II}) for indoor navigation using the 1:8 scaled model car from Fig.~\ref{fig:eb_car}(a) and \textit{III}) on real-world driving with the full scale autonomous driving test vehicle from Fig.~\ref{fig:eb_car}(b), as well as on the nuScenes computer vision dataset.

\subsection{Competing Algorithms and Performance Metrics}

DWA has been implemented based on the Robot Operating System (ROS) DWA local planner, taking into account obstacles provided by the YoloV3 object detector of Redmon and Farhadi~\cite{redmon2018yolov3}. In the case of the PilotNet algorithm proposed by Bojarski et at.~\cite{NVIDIA_end2end_Learning}, the input images are mapped directly to the steering command of the vehicle. The steering commands are executed with an incremental value of $0.01$deg, dependent on the PilotNet's output, while the velocity is controlled using a proportional feedback law, with gain $K=1.6$.

To assess the success rate of each algorithm, the ground truth is considered as the path driven by a human driver. The ground truth of the curvature and road width is calculated as for the trajectory sequences $Z_d$ in the POMDP setup from the Learning a Vision Dynamics Model section. Namely, the curvature is given by the polynomial interpolation of the human driven path, while the road width's ground truth is correlated to the longitudinal velocity of the vehicle.

Ideally, each method should navigate the environment collision-free, at maximum speed and as close as possible to the ground truth. As pointed out by Codevilla et al.~\cite{CodevillaLKD18}, there are limits to the offline policy evaluation employed in Experiments \textit{III}, which can be partially overcome by choosing an appropriate evaluation metric. The cumulative speed-weighted absolute error of Codevilla et al.~\cite{CodevillaLKD18} has been chosen as performance metric. This metric is intended to equally quantify Experiments \textit{I} and \textit{II}, which are pure closed loop experiments, with the offline evaluation performed in Experiments \textit{III}. Additionally, the average speed, the curvature error $e_{c}$ and the processing time have been evaluated. $e_{c}$ represents the difference between the estimated and actual path curvature, calculated using polynomial interpolation, while $e_{xy}$ is defined as:

\begin{equation}
	e_{xy} = \frac{1}{m} \norm{ \sum^{T}_{t=0} (\hat{\vec{p}}^{<i+t>} - \vec{p}^{<i+t>}) \cdot v_{i+t} }_1,
	\label{eq:rmse}
\end{equation}

\noindent where $\hat{\vec{p}}$ and $\vec{p}$ are coordinates on the estimated and human driven trajectories, respectively. $l$ is a lookahead distance calculated for a prediction horizon $\tau_o = 3$s. In order to compare the three competing methods, the metric in Eq.~\ref{eq:rmse} quantifies the total system error.

The percentage of times an algorithm crashed the vehicle and the number of times the destination goals were reached have also been measured for experiments \textit{I} and \textit{II}. In the following, we discuss the obtained values of the computed metrices for the three competing algorithms in the experiments, as summarized in Table~\ref{tab:results}.

\begin{table*}
	\centering
	\begin{tabular}{crllllll}
		\hline
		\multirow{2}{*}{\textbf{Experiments}} & \multirow{2}{*}{\textbf{Method}} & \multirow{2}{*}{\textbf{Crash}} & \textbf{Goal} & \textbf{Avg.} & \multirow{2}{*}{$e_{xy} \pm$ STD [m]} & \multirow{2}{*}{$e_{c} \pm$ STD} & \textbf{Processing} \\
		& & & \textbf{reach} & \textbf{speed} [m/s] & & & \textbf{time} [ms] \\
		\hline
		\textit{I} & DWA-NMPC & \textbf{13}\% & \textbf{87}\% & 4.74 & \textbf{14.03} & \textbf{0.12} & 77 \\
		GridSim & PilotNet & 35\% & 65\% & 5.29 & 30.15 & 0.37 & \textbf{21} \\
		simulation & \textbf{LVD-NMPC} & 15\% & 85\% & \textbf{6.17} & 16.95 & 0.14 & 38 \\
		\hline
		\textit{II} & DWA-NMPC & 30\% & 70\% & 0.83 & 0.21 & 0.26 & 98 \\
		Indoor & PilotNet & 40\% & 60\% & 1.28 & 1.52 & 0.83 & \textbf{53} \\
		navigation & \textbf{LVD-NMPC} & \textbf{20}\% & \textbf{80}\% & \textbf{1.61} & \textbf{0.21} & \textbf{0.20} & 61 \\
		\hline
		\textit{III} & DWA-NMPC & - & - & 25 & 56.38 & 0.14 & 97 \\
		Real-world & PilotNet & - & - & 25 & 118.36 & 0.28 & \textbf{54} \\
		driving & \textbf{LVD-NMPC} & - & - & 25 & \textbf{49.04} & \textbf{0.08} & 63 \\
		\hline
		\textit{III} & DWA-NMPC & - & - & 4.44 & 41.81 & 0.11 & 101 \\
		nuScenes & PilotNet & - & - & 4.44 & 93.50 & 0.26 & \textbf{58} \\
		dataset & \textbf{LVD-NMPC} & - & - & 4.44 & \textbf{37.02} & \textbf{0.07} & 66 \\
		\hline
	\end{tabular}
	\caption{Results for experiments \textit{I}, \textit{II} and \textit{III}, where STD represents the standard deviation.}
	\label{tab:results}
\end{table*}


\subsection{Experiment I: Simulation Algorithm Comparison}
\label{sec:experiments_A}

The first set of experiments are simulations over 10 goal-navigation trials performed in GridSim. GridSim\footnotetext{Additional information available at \url{www.rovislab.com/gridsim.html}.}, proposed by Trasnea et al.~\cite{Trasnea_IRC2019}, is our autonomous driving simulation simulation engine that uses kinematic models to generate synthetic occupancy grids from simulated sensors. It allows for multiple driving scenarios to be easily represented and loaded into the simulator. The simulation parameters are the same as in the NeuroTrajectory state trajectory planning approach of Grigorescu et al.~\cite{Grigorescu_RAL_2019}.

For training PilotNet and LDV-NMPC, the goal navigation task was run over $10$ driving routes, as follows. A trajectories database has been mapped to sensory information, while the ego-vehicle was manually driven by a person. This resulted in over $200.000$ pairs of data samples. Since GridSim provides observations in the form of occupancy grids, the raw input data consisted of distances between the vehicle and the obstacles, sampled at an angle of $2^{\circ}$, without any visual observation being considered. Due to the simplified structure of the occupancy grid observations, which does not consider visual data, the sets of simulated experiments are used as simple sanity checks for evaluating the competing methods.

Overall, as indicated by the values of the percentage of crashes ($13\%$), goal reach ($87\%$) and position ($14.03$m) and curvature ($0.12$) errors from Experiments set I (GridSim simulation) in the first row of Table~\ref{tab:results}, the analytical non-learning DWA-NMPC approach provided top quantitative results in this experiment. This comes to no surprise, since the full state of the vehicle and the precise locations of the obstacles are known a-priori. Nevertheless, as pointed by the results for GridSim simulation if Table~\ref{tab:results}, LVD-NMPC came relatively close to similar values for the computed metrices ($15\%$ crashes, $85\%$ goal reaching, an average speed of $6.17$m/s, $16.95$m and $0.14$ for position and curvature errors, respectively, as well as $38$ms computation time), even outperforming the other methods is terms of vehicle speed. As also shown in the next experiments, PilotNet provided the fastest processing time, mainly due to the fact that the most cost efficient component of PilotNet is the forward propagation of the input data over its neural network. The DaGGer effect could be avoided, since the simulation environment allowed us to gather as much training data as necessary.

\subsection{Experiment II: Indoor Algorithm Comparison}
\label{sec:experiments_B}

In this experiment we have tested using the 1:8 scaled Audi model-car vehicle from Fig.~\ref{fig:eb_car}(a), with different indoor navigation tasks. The reference routes which the car had to follow were defined as straight lines, sinusoids, circles and a $75$m prerecorded loop. The first set of $10$ trials were performed without any obstacles present on the reference routes, while the second $10$ trials set contained static and dynamic obstacles. The static obstacles are composed of cardboard boxes which the model-car can sidestep, while the dynamic obstacles are represented by two other model-cars, manually driven at different speeds and with different trajectories. The state of the vehicle was measured using wheels odometry and an Inertial Measurement Unit (IMU).



LVD-NMPC provided the highest quantitative results, apart from the processing time, which was better for PilotNet. The main reasons for DWA-NMPC's increase in computation time comes from uncertainties in environment perception and localization. This is a common phenomenon encountered in decoupled processing pipelines, where a decrease in perception accuracy produces a decrease in control performance and vice versa. This is not the case with LVD-NMPC, since perception is tightly coupled to motion control through our vision dynamics model. The model-based nature of our algorithm allowed us to outperform a model-free method such as PilotNet, which tends to have a jittering effect in its control output.

\begin{figure}
	\centering
	\begin{center}
		\includegraphics[scale=0.8]{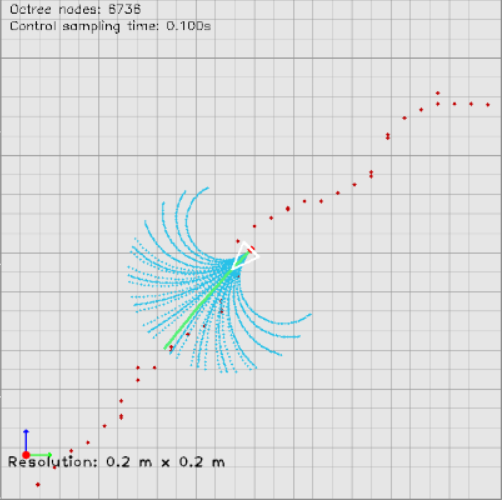}
		\caption{\textbf{Desired trajectory estimation using LVD-NMPC}. The vehicle selects the best desired trajectory (green) from the set of possible candidates (blue) (best viewed in color).}
        \label{fig:candidate_trajectories}
	\end{center}
\end{figure}

A snapshot from the control loop of LVD-NMPC is shown in Fig.~\ref{fig:candidate_trajectories}, where the desired trajectory (depicted in green) is calculated using the output of the proposed deep neural network from Fig~\ref{fig:neural_network_diagram} and a set of candidate trajectories (shown in blue) calculated using the dynamic model of the vehicle from Eq.~\ref{eq:nominal_process_model}. We have observed that the advantage of LVD-NMPC relies in the combination of the analytical dynamic model of the car, which is subsequently adapted to unseen situations by the deep network's estimations, as specified in the state estimation equation~\ref{eq:true_system_model}.

\begin{figure}
	\centering
	\begin{center}
		\includegraphics[scale=1.05]{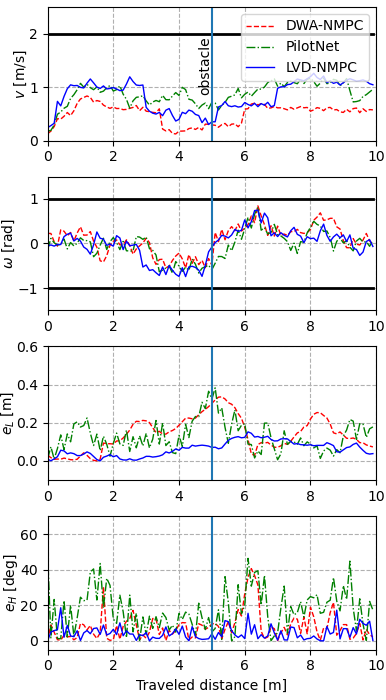}
		\caption{\textbf{Velocity, steering, position errors and heading errors vs. a $10$m travel distance using the model-car from Fig.~\ref{fig:eb_car}(a)}. LVD-NMPC provides a smoother vehicle trajectory, leveraging on learned obstacles and environmental landmarks when learning the vision dynamics model. Actuator constraints are shown with black lines.}
        \label{fig:path_tracking_modelcar_10m}
	\end{center}
\end{figure}

Fig.~\ref{fig:path_tracking_modelcar_10m} illustrates velocity, steering, position errors and heading errors for a trial distance of $10$m, containing an obstacle at position $(0\text{m}, 5\text{m})$ on the reference route. The obstacle's position is relative to the starting position. It can be observed that LVD-NMPC has a smoother trajectory, starting to adapt the control inputs earlier than the competing methods. DWA-NMPC and PilotNet are more aggressive, a jittering effect being encountered in both control outputs. This uncertainty is correlated to the accuracy of the perception system and the lack of a system model in the case of DWA-NMPC and PilotNet, respectively.

In order to assess the behavior of the methods with respect to the DaGGer effect, we have placed on the reference route obstacles which were not given at training time. The DNNs embedded in LVD-NMPC and PilotNet managed to bypass the obstacles, leveraging on environment landmarks present in the training data. This points to the fact that although the learning based approaches were able to correctly adjust the vehicle's trajectory, they still require enough training data to recognize environment landmarks.

\subsection{Experiment III: On-Road Algorithm Comparison}
\label{sec:experiments_C}

Finally, the third experiment tested LVD-NMPC on over $100$km of real-world driving in different environments. 
In total, approx. $463.000$ samples were acquired and varied into $75\%$ training and $25\%$ test sets. This data has been used for training the deep network from Fig.~\ref{fig:neural_network_diagram} in a self-supervised fashion, where pairs of observations (images) and labels (driving trajectories) were given as input to the optimization function in Eq.~\ref{eq:optimal_action_value_function}. Although the training performed in a self-supervised manner, the same real-world data could be used to train a DNN solely based on the observation using Inverse Deep Reinforcement Learning. Such a method was proposed by Wulfmeier et al.~\cite{WulfmeierWP16}, where cost functions for mobile navigation have been learned using Inverse DRL. Nevertheless, the most straightforward approach to self-driving using DRL is to learn the parameters of the network in a simulated environment, where a car would autonomously explore its driving environment. In such a system, the reward function would change the position and orientation of the car based on the visual input. Finally, the challenge in this case would be the mapping of the trained DNN to the real-world vehicle.

In addition to our own real-world driving data, we have evaluated the competing methods on the nuScenes dataset (\url{https://www.nuscenes.org/}). Among different benchmarking datasets, we have chosen nuScenes due to its sensor setup and odometry information. The data collection contains over $15$h of driving data divided into $1000$ driving scenes collected in Boston and Singapore, each scene having a length of $20$s. $242$km were covered at an average speed of $16$km/h. We have converted the ego-vehicle poses to our global 2D reference path coordinates. The original $1600\text{px} \times 900\text{px}$ resolution has been downscaled to $640\text{px} \times 360\text{px}$.

We have encountered similar results to the ones in experiment \textit{II}, with LVD-NMPC providing a more accurate estimate of control outputs. Due to the fact that the test vehicle is a real-sized car (as opposed to the 1:8 Audi model-car), the values of $e_{xy}$ are larger that in experiment \textit{II}. On the other hand, the curvature error is lower, since the driving itself contained less curves than in the case of the indoor navigation experiment. The results on the nuScene dataset were slightly better, mainly due to the relative low speed of the vehicle during data acquisition.


As shown by the results from the real-world experiments \textit{I} and \textit{II}, given in Table~\ref{tab:results}, our model delivers an inference time of slightly above $60$ms on an embedded NVIDIA AGX Xavier development board, equipped with an integrated Volta GPU processor, having a 512 CUDA cores. Depending on the speed of the vehicle, this inference time could be sufficient if the vehicle is traveling at a relatively low speed. However, an increase in computation time is required for high speed vehicles, where the environment also varies with a higher speed.

The metrics used in Table~\ref{tab:results} could be aggregated together in a single metric, where each element, that is percentages of crashes and goal reach, average speed, position and curvature errors and processing time, would be combined in a single weighted function. Nevertheless, in this case the intrinsic values of the individual measurements would be lost. As an example, a model yielding an optimal metric value due to high processing time could crash more often than a model which is slower in terms of computation time.

\section{Conclusions}
\label{sec:conclusions}

This paper introduces the \textit{learning-based vision dynamics nonlinear model predictive control} approach for controlling autonomous vehicles. The method uses a deep neural network as a vision dynamics model which estimates both the desired state trajectory of the vehicle, given as input to a constrained non-linear model predictive controller, as well as the weighting gains of the aforementioned controller. One of the advantages of LVD-NMPC is that the Q-learning training is self-supervised, without requiring the manual annotation of the training data. The experimental results show the robustness of the approach with respect to state-of-the-art competing algorithms, both classical, as well as learning based.

As future work, we plan to investigate the stability of LVD-NMPC, especially in relation to the functional safety requirements needed for automotive grade deployment. Being already implemented on an embedded device, that is the NVIDIA AGX Xavier, we believe that the controller can be used on real-world cars, provided that the safety requirements are meet. Setting safety aside, its current implementation is directly linked to the computation power of the vehicle's computer. The faster its deep learning accelerator is, the more dynamic situations LVD-NMPC could cope with.

\begin{acks}
This work has received funding from the European Union's Horizon 2020 research and innovation programme under grant agreement No 800928, European Processor Initiative EPI.
\end{acks}

\bibliographystyle{SageV}
\bibliography{references}

\end{document}